%% file: main.tex
\documentclass[10pt,twocolumn,letterpaper]{article}

\usepackage{cvpr}              % To produce the CAMERA-READY version
\usepackage[accsupp]{axessibility}  % Improves PDF readability for those with disabilities.

\input{preamble}
\definecolor{cvprblue}{rgb}{0.21,0.49,0.74}
\usepackage[pagebackref,breaklinks,colorlinks,allcolors=cvprblue]{hyperref}
\usepackage{marvosym}
\def\paperID{Null} % *** Enter the Paper ID here
\def\confName{CVPR}
\def\confYear{2026}

\title{\ours: Advancing GUI Agents with Effective Reinforcement Learning and Precise Inference-Time Grounding}

\author{
Shuquan Lian$^1$ \quad
Yuhang Wu$^1$ \quad
Jia Ma$^1$ \quad
Yifan Ding$^1$ \quad
Zihan Song$^1$ \quad
Bingqi Chen$^1$ \quad \\
Xiawu Zheng$^1$ \quad
Hui Li$^{1}$\textsuperscript{\Letter} \quad
Rongrong Ji$^{1,2}$ \\
\\
$^1$Key Laboratory of Multimedia Trusted Perception and Efficient Computing Ministry of \\
Education of China, Xiamen University \\
$^2$Sino-Russian Research Center for Digital Economy \\
{\tt\small shuquanlian@stu.xmu.edu.cn, hui@xmu.edu.cn}
}
\usepackage{algorithm}
\usepackage{algorithmic}
\usepackage{amsmath}
\usepackage{booktabs}
\usepackage{multirow}
\usepackage{colortbl}
\usepackage[table]{xcolor}
\usepackage{xspace}
\usepackage{enumitem}

\newcommand{\hideit}[1]{}

\newcommand{\ours}{UI-AGILE\xspace}

\begin{document}

\maketitle
\vspace{-30pt}
\begin{abstract}
The emergence of Multimodal Large Language Models has driven significant advances in Graphical User Interface (GUI) agent capabilities. Nevertheless, we recognize that existing training and inference techniques for GUI agents still suffer from a dilemma for reasoning designs, ineffective reward, and visual noise. To address these issues, we introduce \ours for enhancing GUI agents at both training and inference. For training, we propose a suite of improvements to the Reinforcement Fine-Tuning (RFT) process: 1) a ``Simple Thinking'' strategy to balance planning with speed and grounding accuracy, 2) a continuous reward function to incentivize high-precision grounding, and 3) a cropping-based resampling strategy to mitigate the sparse reward problem and improve learning on complex tasks. For inference, we present decomposed grounding with selection to improve grounding accuracy on high-resolution displays by breaking the image into smaller, manageable parts. Experiments show that \ours exhibits better grounding performance than state-of-the-art methods on benchmarks ScreenSpot-Pro and ScreenSpot-v2, while showing strong agent capabilities. For instance, using both our training and inference enhancements brings 23\% grounding accuracy improvement over the best baseline on ScreenSpot-Pro. The implementation of \ours is provided at \url{https://github.com/KDEGroup/UI-AGILE}.

\end{abstract}
\vspace{-20pt}

\input{tex/introduction} 
\input{tex/preliminary_analysis}
\input{tex/method}

\input{tex/experiment}
\input{tex/related_work}

\input{tex/conclusion}

\section{Acknowledgment}
This work is supported by the National Key Research and Development Program of China (No. 2025YFE0113500), the National Science Fund for Distinguished Young Scholars (No. 62525605), and the National Natural Science Foundation of China (No. 62572410, No. 62272401, and No. U22B2051).

{
    \small
    \bibliographystyle{ieeenat_fullname}
    \bibliography{main}
}

\input{tex/X_suppl}

\end{document}

%% file: tex/introduction.tex
\section{Introduction}
\label{sec:intro}

%  in image understanding~\citep{abs-2502-13923}

Driven by the growing capabilities of Multimodal Large Language Models, Graphical User Interface (GUI) agents, which execute tasks by understanding screenshots and user instructions, are advancing rapidly~\citep{ZhangHQ0LQ0MLLR25}.

Prior methods for GUI agents mostly rely on Supervised Fine-Tuning (SFT), requiring a large amount of human-annotated or synthesized data for teaching agents how to plan its actions and grounding~\citep{GouWZXCS0025, 0003WXWSJC0CL025, LinLGYWBLWS25}. 
% 省略一些~\citep{abs-2501-12326, ChengSCX0Z024, GouWZXCS0025, 0003WXWSJC0CL025, abs-2412-04454, LinLGYWBLWS25}. 
Recently, Reinforcement Fine-Tuning (RFT) has been proposed to enhance GUI agents~\citep{abs-2503-21620, abs-2504-10458}, and achieves considerable improvements compared to previous approaches.
% ~\citep{chen2025r1v, abs-2503-07523, abs-2503-01785, abs-2503-07536, abs-2503-07365}
% 省略一些 ~\citep{ abs-2503-01785, abs-2503-07536, abs-2503-07365}. 
% For instance, UI-R1~\citep{abs-2503-21620} and GUI-R1~\citep{abs-2504-10458} apply RFT to enhance GUI agents and achieve considerable improvements compared to previous approaches.

Despite the significant momentum of GUI agent techniques, we recognize that their practical application is hindered by several limitations in both training and inference stages (see Sec.~\ref{sec:preliminary_analysis} for details):
\begin{itemize}

\item \textbf{P1: A Dilemma for Reasoning Designs}: An elaborate reasoning process not only degrades grounding accuracy but also significantly increases training time and inference latency, while a ``No Thinking'' approach exhibits low accuracy for predicting non-grounding actions~\citep{abs-2503-04472}.

\item \textbf{P2: Ineffective Reward}: GUI agents often get stuck on complex interfaces and receive no effective learning signal (i.e., sparse reward). Besides, simple binary feedback (correct/incorrect), a design used by many existing methods~\citep{abs-2503-21620, abs-2504-10458} may fail to endow agents with the ability to perform precise localization.

\item \textbf{P3: Visual Noise}: Even well-trained agents frequently struggle to cope with high-resolution screens, as irrelevant visual noise degrades their grounding accuracy.
\end{itemize}

To address the above problems, we propose \ours, a framework aiming at improving both the RFT and the inference stages of GUI agents.
Its contributions are as follows:
\begin{itemize}

% (Sec.~\ref{sec:simple_thinking})
\item To overcome \textbf{P1}, \ours applies a ``Simple Thinking'' strategy that employs reasoning with appropriate lengths when trained on GUI grounding data, while not restricting the reasoning length when trained on data for other actions. ``Simple Thinking'' effectively reduces training and inference cost while also balancing both the core grounding task and the prediction of non-grounding actions.
% \ours operationalize ``Simple Thinking'' through a specialized reward function.  

% (Sec.~\ref{sec:continuous_reward})
% (Sec.~\ref{sec:crop})
\item To tackle \textbf{P2}, \ours employs cropping-based resampling to dynamically adjust the difficulty of training samples to avoid ineffective training with zero reward. Furthermore, \ours harnesses a design of continuous grounding reward for the RFT stage to incentivize more precise localization to the target's center. 
% instead of using the common binary reward

% (Sec.~\ref{sec:decomposed})
\item To solve \textbf{P3}, \ours uses a visual noise reduction method termed decomposed grounding with selection. It decomposes a high-resolution screenshot into multiple sub-images, generates candidate elements on each, and finally uses a  Vision-Language Model (VLM) to ``adjudicate'' the best match. This approach significantly improves the agent's grounding accuracy on high-resolution displays during inference.

\end{itemize}

Extensive experiments validate the effectiveness of our methods. 
Trained on about only 9k samples for just 2 epochs, based on Qwen2.5-VL~\citep{abs-2502-13923}, \ours shows superior grounding performance, while also showcasing strong general agent capabilities. 
Furthermore, our inference method can act as a plug-and-play enhancement for a wide range of existing agents, improving the accuracy of some open-source models.

\hideit{
Overall, on two benchmarks ScreenSpot-Pro and ScreenSpot-v2, our methods achieve the state-of-the-art performance.
For instance, using both our proposed training and inference enhancement methods brings 23\% grounding accuracy improvement over the best baseline on ScreenSpot-Pro. 
}

%% file: tex/preliminary_analysis.tex
\section{Preliminary Analysis}
\label{sec:preliminary_analysis}

To substantiate the problems (\textbf{P1-P3}) outlined in Sec.~\ref{sec:intro}, we conduct a series of pilot studies with the same settings as our experiments in Sec.~\ref{sec:exp}.

\hideit{These preliminary experiments validate our motivations for developing the UI-AGILE framework.}

\subsection{Dilemma of Reasoning (P1)}

We investigate the trade-off between reasoning complexity, grounding accuracy, and agent training time. A preliminary analysis, detailed further in our ablation study (Fig.~\ref{figure:ablation} in Sec.~\ref{sec:exp_ablation}), reveals a clear dilemma: 
\begin{itemize}
    \item ``No Thinking'' models (which directly output actions) achieve slightly better grounding accuracy but suffer from significant performance degradation on general agent abilities.
    
    \item ``Normal Thinking'' models (using elaborate reasoning) achieve better agent abilities but at the cost of degraded grounding accuracy and significantly increased training time (requiring 2x more time than ``No Thinking'' approach). 
    
\end{itemize}

The observation confirms that both of the two strategies adopted by existing works have their limitations, motivating us to adopt ``Simple Thinking'' strategy to balance grounding, planning, and efficiency.

\subsection{Ineffective Reward (P2)}

To quantify the ineffective reward problem, we analyze the normal training process of the GUI agent with RFT. 
We track the frequency of ineffective steps, where all responses generated for a training sample receive zero reward, providing no learning signal.

During the first training epoch, only 61.8\% of training steps were successful on the first attempt. This implies that, without intervention, 19.1\% to 38.2\% training samples (here we have 2 training samples per step) will provide no learning signal. The observation highlights the critical need for a mechanism to mitigate sparse rewards, motivating us to design the cropping-based resampling strategy.

\subsection{Visual Noise (P3)}

Modern electronic devices feature high-resolution displays (e.g., 3840x2160), which, when converted into tokens for a VLM, can result in an overwhelmingly long sequence (e.g., over 10,000 tokens). 
We hypothesize that a significant portion of these tokens represent irrelevant background information, i.e., visual noise.

% , acting as noise that can distract the GUI agent and degrade its grounding accuracy

To validate this hypothesis, we conduct a preliminary experiment on ScreenSpot-Pro~\citep{abs-2504-07981}.
We apply our cropping method that will be introduced  in Sec.~\ref{sec:crop}, but for the purpose of creating a controlled test environment.  
For each original screenshot, we crop it to 1024x1024, ensuring the ground-truth bounding box is contained within the frame.
On this new dataset, the grounding accuracy of UGround-V1-7B~\citep{GouWZXCS0025} shows a significant improvement from 31.6 to 56.0, verifying our hypothesis.

%% file: tex/method.tex
\section{Our Framework \ours}
\label{sec:method}

\begin{figure*}[t]
\centering
\includegraphics[width=0.97\textwidth]{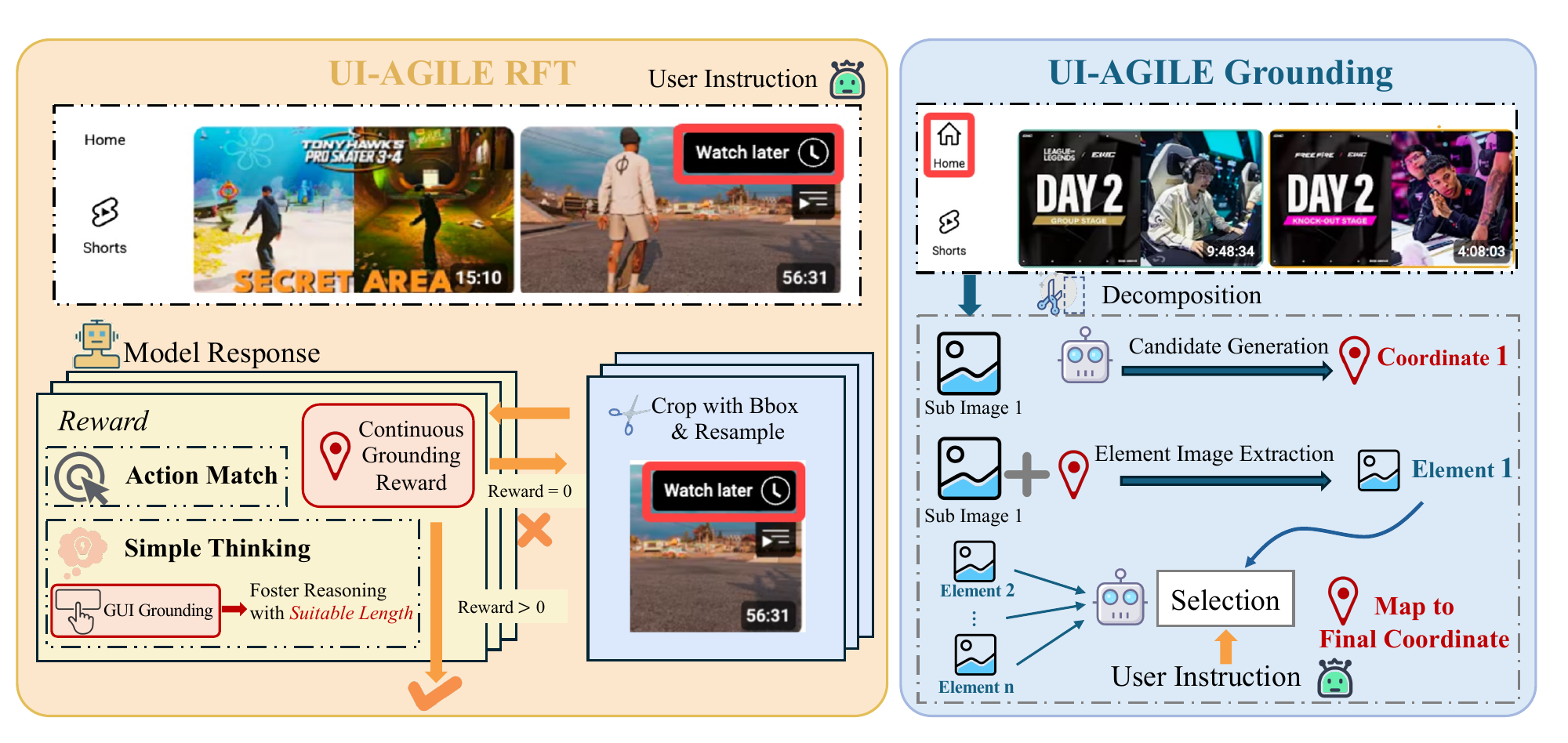} % Reduce the figure size so that it is slightly narrower than the column.
\vspace{-10pt}
\caption{
    An overview of \ours. 
    (1) Left: The training stage is enhanced with our three core contributions: ``Simple Thinking'', continuous grounding reward and cropping-based resampling. Continuous grounding reward being zero would result in crop-based resampling.
    (2) Right: The inference stage uses our proposed decomposed grounding with selection.
}
\vspace{-10pt}
\label{fig:UI_Crop}
\end{figure*}
% , illustrating its training and inference pipelines

In this section, we will illustrate \ours. 
Fig.~\ref{fig:UI_Crop} provides an overview of \ours, which aims at improving both training and inference stages of GUI agents.

\hideit{At training time, \ours adopts a new design for the reward function, which consists of a ``Simple Thinking'' reward for efficient reasoning (Sec.~\ref{sec:simple_thinking}) and a continuous grounding reward (Sec.~\ref{sec:continuous_reward}) for precise localization.
\ours further uses cropping-based resampling (Sec.~\ref{sec:crop}), a novel strategy designed to overcome the sparse reward issue. 
The model generates multiple responses from an image and instruction, which are evaluated by reward functions, including ``Simple Thinking'' reward and continuous grounding reward. 
If a training sample proves too difficult (i.e., receive a grounding reward of zero for all generated responses), the image will be cropped to simplify the task, and the model resamples new responses from this modified input.}

During training, it generates multiple responses from an image and instruction, which are evaluated by reward functions, including ``Simple Thinking'' reward for efficient reasoning (Sec.~\ref{sec:simple_thinking}) and continuous grounding reward (Sec.~\ref{sec:continuous_reward}) for precise localization. 
If a training sample proves too difficult (i.e., receive a grounding reward of zero for all generated responses), the image will be cropped to simplify the task, and the model resamples new responses from this modified input (Sec.~\ref{sec:crop}).

During inference, it applies decomposed grounding with selection (Sec.~\ref{sec:decomposed}) to enhance grounding on the high-resolution displays common in modern applications, making GUI agents more practical for real-world use.

\subsection{``Simple Thinking'' for Reconciling the Reasoning Dilemma (P1)}
\label{sec:simple_thinking}

\hideit{
GRPO~\citep{abs-2402-03300} is widely recognized as a powerful technique for instilling complex reasoning abilities in LLMs, often through rewarding elaborate chains of thought. }

\citet{abs-2503-21620} posit that excessive reasoning is not essential for GUI grounding and can even be detrimental. In addition, excessive reasoning can significantly increase training time and inference time.
However, \emph{the complete role of GUI agents extends beyond mere grounding}, and deciding next action (e.g., click or type) inherently demands a foundational level of reasoning. 

To reconcile the dilemma of whether to apply reasoning or discard excessive reasoning, we propose ``Simple Thinking''. 
It encourages thoughts with an appropriate length when trained with GUI grounding data, operationalized through a specialized reward function. When trained on data for other actions, it does not restrict thought length.

The reward $R_{think}$ is defined as:
\begin{equation}
R_{\text{think}} = I(R_{\text{grounding}}>0) \cdot (R_{\text{length}}(L) + R_{\text{bonus}})
\end{equation}
\begin{align}
\label{eq:simple_thinking_reward}
R_{\text{length}}(L) &= 
\begin{cases} 
    1.0 & \text{if } l_{\text{ideal\_start}} < L \le l_{\text{ideal\_end}} \\
    F1 & \text{if } l_{\text{min}} < L \le l_{\text{ideal\_start}} \\
    F2 & \text{if } l_{\text{ideal\_end}} < L < l_{\text{max}} \\
    0 & \text{otherwise}
\end{cases} \\
F1 &= \frac{1}{2} \left(1 - \cos\left(\pi \frac{L - l_{\text{min}}}{l_{\text{ideal\_start}} - l_{\text{min}}}\right)\right) \\
F2 &= \frac{1}{2} \left(1 + \cos\left(\pi \frac{L - l_{\text{ideal\_end}}}{l_{\text{max}} - l_{\text{ideal\_end}}}\right)\right)
\end{align}
where:
\begin{itemize}
    \item $I(R_{\text{grounding}}>0)$ is an indicator function that grants reward only when the grounding reward $R_{\text{grounding}} > 0$, linking reasoning to effective outcomes.
    
    \item $R_{\text{length}}(L)$ is a non-linear reward based on the reasoning length $L$. $l_{\text{ideal\_start}}$ and $l_{\text{ideal\_end}}$ define an ideal range of reasoning length, where reward is 1. The reward will be zero if the reasoning length exceeds $l_{\text{min}}$ or $l_{\text{max}}$.
    
    \item $R_{\text{bonus}}$ is a fixed bonus for syntactically complete thoughts (e.g., ending with proper punctuation), encouraging structured reasoning.
\end{itemize}

This function defines an ideal range where the reward is maximized, encouraging thoughts that are neither too brief (``under-thinking'') nor too verbose (``over-thinking'').
Outside this ideal range, it uses the cosine function for smooth degradation down to a reward of zero at the absolute bounds. 
This smooth, non-linear penalty, inspired by the Hann window~\citep{Oppenheim09}, a concept fundamental to signal processing, provides a more stable learning signal for RL than a hard cliff.
Furthermore, the additional bonus for syntactically complete thoughts discourages incomplete reasoning, thereby ensuring better training stability.

\subsection{Continuous Grounding Reward for Precise Localization (P2)}
\label{sec:continuous_reward}

\hideit{Prior works~\citep{abs-2503-21620, abs-2504-10458} shift the focus of GUI agent evaluation from traditional object grounding (i.e., IoU) to the precision of the action coordinate. 
To this end, they typically employ a simple binary reward for localization: a reward of 1 for a correct prediction (e.g., inside the target radius) and 0 otherwise.}

Unlike general visual grounding tasks which compute the IoU between the predicted bounding box and the ground truth box, GUI agents usually predict a click point and receive a simple binary reward in rule-based RFT~\citep{abs-2503-21620, abs-2504-10458}: a reward of 1 for a correct prediction (e.g., inside the ground truth box) and 0 otherwise. 
However, this binary reward is insufficient for high-precision control, as it equally rewards clicks on an element's edge and at its center. This non-discriminatory feedback misguides the model to learn an element's boundaries rather than its semantic core.

To resolve this issue, we introduce a continuous grounding reward. 
It is calculated as a function of the distance from the predicted point to the center of the ground-truth bounding box:
\begin{equation}
\label{eq:continuous_reward}
R(x, y) = 
\begin{cases} 
    1 + \exp(-C \cdot d_{\text{norm}}^2) & \text{if } (x, y) \in \text{BBox} \\
    0 & \text{otherwise}
\end{cases}
\end{equation}
where:
\begin{itemize}
    \item $R(x, y)$ is the reward score for the predicted coordinate.
    
    \item $(x, y)$ is the coordinate predicted by the agent.
    
    \item $\text{BBox}$ is the ground-truth bounding box, defined by its top-left $(x_1, y_1)$ and bottom-right $(x_2, y_2)$ corners.

    \item $C$ is a coefficient that affects the smoothness of the curve.
    
    \item $d_{\text{norm}}$ is the Chebyshev distance (or $L_\infty$ norm) of the point from the center of the bounding box, normalized by the box's dimensions. It is calculated as:
    \begin{equation}
    \label{eq:norm_dist}
    d_{\text{norm}} = \max\left(\frac{\left|x - c_x\right|}{w_h}, \frac{\left|y - c_y\right|}{h_h}\right)
    \end{equation}
    Here, $(c_x, c_y) = (\frac{x_1+x_2}{2}, \frac{y_1+y_2}{2})$ represents the center of the bounding box, and $(w_h, h_h) = (\frac{x_2-x_1}{2}, \frac{y_2-y_1}{2})$ are its half-width and half-height, respectively.
    
\end{itemize}

We employ the Chebyshev distance instead of the Euclidean distance because the reward contours generated by the Chebyshev distance are squares, which geometrically align with the rectangular shape of GUI bounding boxes. 
The exponential term $\exp(-C \cdot d_{\text{norm}}^2)$ provides a Gaussian-like reward landscape~\citep{Oppenheim09} with a peak at the center. This creates a strong, differentiable gradient signal that guides the agent precisely to the element's semantic core rather than its boundaries.

\subsection{Cropping-Based Resampling for Sparse Reward Mitigation (P2)}
\label{sec:crop}

During GRPO training, GUI agents often face the sparse reward challenge, particularly on complex tasks. 
When the model fails to place its prediction within the correct bounding box for a given screenshot over all generations, it receives no positive signals, leading to training stagnation, thus difficult samples cannot contribute to model improvement.

Inspired by curriculum learning~\citep{BengioLCW09}, which presents training examples in an easy-to-hard progression, we propose Cropping-based Resampling. This method acts as a dynamic difficulty adjustment mechanism: if a sample yields zero reward over all generations, we deem it too difficult and reduce its complexity by cropping the screenshot. The resulting smaller crop is guaranteed to fully contain the ground-truth bounding box.

\begin{figure*}[t]
\centering
\includegraphics[width=1\textwidth]{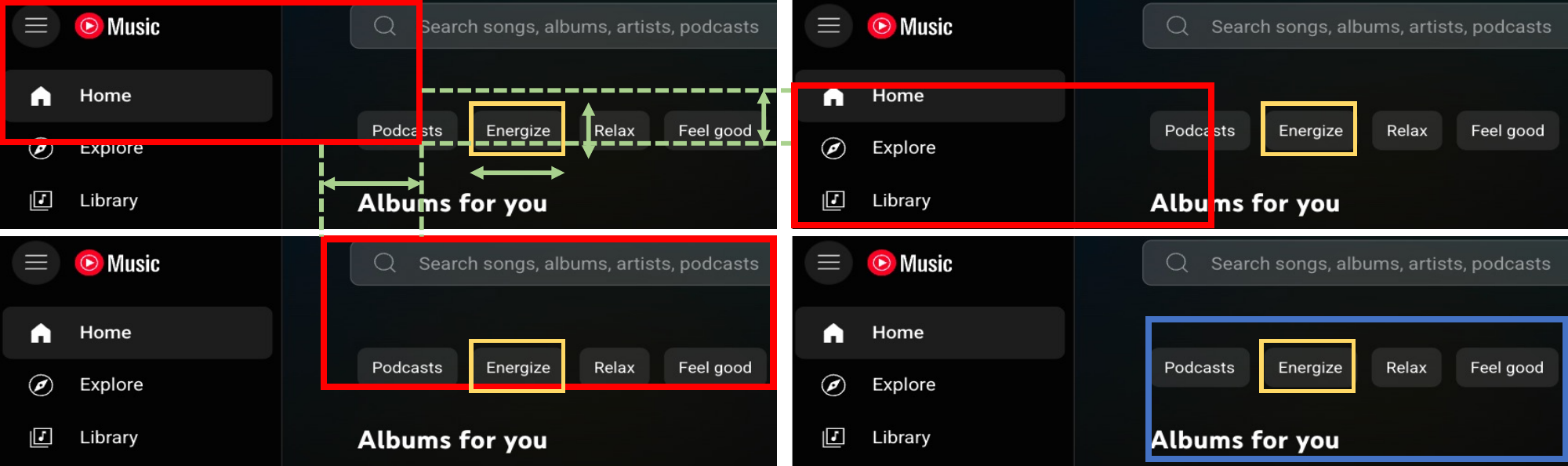}
\vspace{-20pt}
\caption{An example of cropping-based resampling. Yellow bounding boxes are the ground truth, red bounding boxes are invalid cropping, and blue bounding boxes are valid cropping. Green arrows show that the overlap of cropping windows is equivalent to the width or height of the ground-truth bounding box.}
\vspace{-10pt}
\label{crop_with_bbox}
\end{figure*}

A naive implementation is to center the ground-truth bounding box (bbox) in the new cropping, but the model would learn a trivial shortcut, such as developing a bias for predicting the image center. 
We opt to employ a scanning approach as illustrated in Alg.~\ref{alg:algorithm}, ensuring that the cropped image fully contains the ground-truth bounding box. 
It firstly determines the size of cropping based on a predefined ratio (lines 1-2).
Then, the horizontal stride $step_x$ is set to the difference between the cropping width and the bounding box width, while the vertical stride $step_y$ is set to the difference between their respective heights (lines 3-5).
After that, it iterates through all possible cropping windows from left-to-right and top-to-bottom across the original screenshot with the horizontal stride and the vertical stride (lines 7-18).
A random one of windows that fully contains the ground-truth bbox is selected as the new, resampled input (lines 12-16). 
Fig.~\ref{crop_with_bbox} illustrates how our scanning approach identifies valid cropping windows that fully contain the ground-truth bounding box.

Cropping-based resampling dynamically simplifies difficult samples to ensure that they are learnable, allowing the model to leverage more data in fewer epochs, yielding superior results within a similar amount of training time.

\subsection{Decomposed Grounding with Selection for Visual Noise Reduction (P3)}
\label{sec:decomposed}

% Building on the principle of task simplification via cropping from the training stage (Section~\ref{sec:crop}), we now extend this core idea to address a critical bottleneck at inference time, where even well-trained GUI agents often exhibit a sharp decline in performance when deployed in high-resolution environments that challenge their ability to discern fine-grained details from a noisy background.

\begin{algorithm}[t]
\caption{Cropping-Based Resampling}
\footnotesize
\label{alg:algorithm}
\textbf{Input}: $Image$, $Bbox$, $scalingFactor f$\\
\textbf{Output}: $Set(Image_{crop}, Bbox_{crop})$
\begin{algorithmic}[1]
\STATE $(w_o, h_o) \leftarrow GetWidthAndHeight(Image)$
\STATE $(w_{crop}, h_{crop}) \leftarrow (w_o \times f, h_o \times f)$
\STATE $(w_b, h_b) \leftarrow (Bbox[2] - Bbox[0], Bbox[3] - Bbox[1])$
\STATE $step_x \leftarrow w_{crop} - w_b$
\STATE $step_y \leftarrow h_{crop} - h_b$
\STATE $T \leftarrow \emptyset$
\FOR{$xcrop_{min}$ from 0 to $w_o$ step=$step_x$}
\STATE $xcrop_{max} \leftarrow min(xcrop_{min} + w_{crop}, w_o)$
\FOR{$ycrop_{min}$ from 0 to $h_o$ step=$step_y$}
\STATE $ycrop_{min} \leftarrow min(ycrop_{min} + h_{crop}, h_o)$
\STATE \parbox[t]{\dimexpr\linewidth-\algorithmicindent}{$Coord_{crop} \leftarrow [xcrop_{min}, \\
    \hspace*{\algorithmicindent} ycrop_{min}, xcrop_{max}, ycrop_{max}]$
}
\IF{$Bbox$ is contained within $Coord_{crop}$}
\STATE $Image_{crop} \leftarrow Crop(Image, Coord_{crop})$
\STATE $Bbox_{crop} \leftarrow Bbox - Coord_{crop}$
\STATE $Set \leftarrow Set \cup {(Image_{crop}, Bbox_{crop})}$
\ENDIF
\ENDFOR
\ENDFOR
\STATE \textbf{return} $Set$
\end{algorithmic}
\end{algorithm}

It is straightforward to consider applying the cropping-based method to alleviate the visual noise.
However, during inference, the location of the ground-truth bounding box is unknown, making such oracle-based cropping impossible. 
To solve this issue, we propose a multi-stage decomposed grounding with selection method as shown in the right part of Fig.~\ref{fig:UI_Crop}. 
It reduces visual noise by cropping an image into several sub-images and predicts the coordinate individually, while trying to maintain full ground-truth bounding box area. 
The detailed process is as follows:
\begin{enumerate}[label=\textbf{\arabic*.}]
\item \textbf{Decomposition:} The input screenshot is divided into several overlapping sub-images, breaking down the high-resolution screen into smaller, manageable regions.

\item \textbf{Candidate Generation:} The model performs grounding independently on each sub-image to get coordinates, which serve as candidate points.

\item \textbf{Element Image Extraction:} For each candidate point, we extract the corresponding element image by cropping a bounding box centered on the candidate point from the sub-image.

\item \textbf{Selection:} %We prompt the VLM with the user's instruction, the candidate element image, and a direct question asking whether the image match the instruction as Appendix~B details. Then use the model's output logit for the ``Yes'' token as a direct relevance score for the candidate. The candidate with the highest ``Yes'' score is chosen as the final answer, and its corresponding coordinates are remapped to the original screenshot. This QA-based scoring allows the model to perform deeper contextual prediction. 
We use a VLM to score each candidate image based on a direct ``Yes/No'' question asking if it matches the user's instruction (see Appendix~\ref{sec:env_impl_details} for the prompt). The output logit for ``Yes'' serves as the relevance score. The coordinates of the highest-scoring candidate are then remapped to the original screenshot. This QA-based process enables deeper contextual prediction. 
% We also tried another way to select,  analyzed in Sec.~\ref{sec:selection_analysis}.

\end{enumerate}

This process directly benefits from the continuous reward function, as it trains the model to predict points closer to the center of target elements, leading to higher-quality extracted images and thus a more accurate final selection.

\vspace{5pt}
\noindent\textbf{Analysis of Inference Cost.} 
We analyze the inference latency of decomposed grounding with selection by breaking it down into prefilling, decoding and selection stages. 

Counter-intuitively, our approach can \emph{theoretically accelerate the computationally-heavy prefilling stage}. 
Recall that the self-attention mechanism has a quadratic time complexity $\mathcal{O}(n^2)$ concerning the input sequence length $n$~\citep{Vae-aaniSPUJGKP17}. 
By splitting a single large image with $n$ tokens into 4 sub-images (each with roughly n/4 tokens), the total computational cost for attention scales proportionally to $ 4\times (\frac{n}{4} )^2 = \frac{n^2}{4}$, suggesting a theoretical speedup. 
Such cost reduction far outweighs the slightly increased overhead of repeatedly processing the text prompt for each sub-image.

While the decoding process runs for each sub-image, the small number of output tokens (applying ``Simple Thinking'') per run ensures that the cumulative decoding cost will not increase much. 

Finally, the VLM-based selection stage is computationally inexpensive. 
The input element images are very small, and the process only requires a single forward pass to acquire the logits for a ``Yes/No'' answer.

Overall, the cost of applying decomposed grounding with selection is low. 
In addition to the above analysis, we provide results on actual running time in Appendix~\ref{sec:inference_time}. 
Crucially, we believe this overhead could be eliminated or even reversed with future optimizations in inference engines tailored for this ``many small requests'' workload.
%, potentially making our high-accuracy method faster than the baseline.

%% file: tex/experiment.tex
\section{Experiment}
\label{sec:exp}

\subsection{Implementation Details}

% related to GUI tasks 

\vspace{5pt}
\noindent\textbf{Data.} We collect data from multiple open-source datasets, including UI-R1~\citep{abs-2503-21620}, GUI-R1~\citep{abs-2504-10458}, Aguvis~\citep{abs-2412-04454} and Grounding-R1~\citep{yang2025groundingr1}. We filter them using OmniParser~\citep{WanSYLC0BY024} following Grounding-R1~\citep{yang2025groundingr1}. 
We randomly sample 9k examples to train \ours. The value of hyperparameters in Eq.~\ref{eq:simple_thinking_reward} and Eq.~\ref{eq:continuous_reward} are chosen by grid search.

\vspace{5pt}
\noindent\textbf{Environment, Implementation and Hyperparameters.}
Due to space limit, we include the details in Appendix~\ref{sec:env_impl_details} and Appendix~\ref{sec:code}.

\subsection{Grounding Capability Evaluation}

\begin{table*}[!ht]
\centering
\footnotesize
\setlength{\tabcolsep}{3pt}
\caption{Grounding accuracy on ScreenSpot-Pro. ``+Decomposed Grounding'' denotes that the model uses decomposed grounding with selection for enhancing inference. Bold and underlined Results represent the best performance and the second-best performance, respectively. pass@4 indicates the success rate where a task is considered solved if the prediction of at least one of the sub-images is correct.}
\vspace{-10pt}
\begin{tabular}{lcccccccccccccc|cc}
\toprule
\multirow{2}{*}{\textbf{Model}} & \multirow{2}{*}{\textbf{Examples}} & \multirow{2}{*}{\textbf{Epochs}} & \multicolumn{2}{c}{\textbf{Dev}} & \multicolumn{2}{c}{\textbf{Creative}} & \multicolumn{2}{c}{\textbf{CAD}} & \multicolumn{2}{c}{\textbf{Scientific}} & \multicolumn{2}{c}{\textbf{Office}} & \multicolumn{2}{c}{\textbf{OS}} & \multirow{2}{*}{\textbf{Avg}} & \multirow{2}{*}{\textbf{pass@4}} \\
\cmidrule(lr){4-5} \cmidrule(lr){6-7} \cmidrule(lr){8-9} \cmidrule(lr){10-11} \cmidrule(lr){12-13} \cmidrule(lr){14-15}
& & & \textbf{Text} & \textbf{Icon} & \textbf{Text} & \textbf{Icon} & \textbf{Text} & \textbf{Icon} & \textbf{Text} & \textbf{Icon} & \textbf{Text} & \textbf{Icon} & \textbf{Text} & \textbf{Icon} & \\
\midrule
\multicolumn{15}{l}{\textbf{Supervised Fine-tuning}} \\

\midrule
CogAgent-18B & 222M & - & 14.9 & 0.7 & 9.6 & 0.0 & 7.1 & 3.1 & 22.2 & 1.8 & 13.0 & 0.0 & 5.6 & 0.0 & 7.7 & - \\
Aria-UI & 16.6M & - & 16.2 & 0.0 & 23.7 & 2.1 & 7.6 & 1.6 & 27.1 & 6.4 & 20.3 & 1.9 & 4.7 & 0.0 & 11.3 & - \\
ShowUI-2B & 256K & - & 16.9 & 1.4 & 9.1 & 0.0 & 2.5 & 0.0 & 13.2 & 7.3 & 15.3 & 7.5 & 10.3 & 2.2 & 7.7 & - \\
JEDI-3B & 4M & - & 61.0 & 13.8 & 53.5 & 8.4 & 27.4 & 9.4 & 54.2 & 18.2 & 64.4 & 32.1 & 38.3 & 9.0 & 36.1 & - \\
JEDI-7B & 4M & - & 42.9 & 11.0 & 50.0 & \underline{11.9} & 38.0 & 14.1 & \textbf{72.9} & 25.5 & \underline{75.1} & \textbf{47.2} & 33.6 & 16.9 & 39.5 & - \\
\midrule
OS-Atlas-7B & 13M & - & 33.8 & 1.4 & 30.8 & 3.5 & 12.2 & 3.1 & 33.3 & 9.1 & 33.3 & 9.4 & 26.2 & 3.4 & 18.9 & - \\ \rowcolor{gray!35}  % set next line gray background color
+Decomposed Grounding & - & - & 49.4 & 5.5 & 52.0 & 5.6 & 26.4 & 6.3 & 54.9 & 18.2 & 57.6 & 18.9 & 49.5 & 9.0 & 33.1 & 42.1 \\
\midrule
Aguvis-7B & 1M & 1 & 30.5 & 0.7 & 28.8 & 2.8 & 14.7 & 1.6 & 45.8 & 8.2 & 38.4 & 11.3 & 30.8 & 2.3 & 20.4 & - \\ \rowcolor{gray!35}
+Decomposed Grounding & - & - & 50.6 & 11.7 & \underline{60.1} & 7.0 & 31.0 & 4.7 & 62.5 & 20.0 & 63.3 & 18.9 & 45.8 & 6.7 & 36.5 & 44.3 \\
\cmidrule(lr){1-15}
UGround-V1-7B & 10M & - & 51.3 & 5.5 & 48.5 & 8.3 & 18.8 & 1.6 & 59.7 & 14.6 & 59.9 & 17.0 & 40.2 & 7.9 & 31.6 & - \\ \rowcolor{gray!35}
+Decomposed Grounding & - & - & 57.8 & 14.5 & 49.0 & \underline{11.9} & 20.3 & 7.8 & 62.5 & 21.8 & 67.8 & 18.9 & 48.6 & 14.6 & 36.6 & 47.3 \\
\midrule
UI-TARS-2B & - & - & 47.7 & 4.1 & 42.9 & 6.3 & 17.8 & 4.7 & 56.9 & 17.3 & 50.3 & 17.0 & 21.5 & 5.6 & 27.7 & - \\
UI-TARS-7B & - & - & 58.4 & 12.4 & 50.0 & 9.1 & 20.8 & 9.4 & 63.9 & \textbf{31.8} & 63.3 & 20.8 & 30.8 & 16.9 & 35.7 & - \\ \rowcolor{gray!35}
+Decomposed Grounding & - & - & 59.7 & \underline{19.3} & 54.0 & \textbf{15.4} & 38.1 & 12.5 & 63.2 & 27.3 & 71.8 & 28.3 & 45.8 & \underline{21.3} & 41.9 & 50.3 \\
UI-TARS-72B & - & - & 63.0 & 17.3 & 57.1 & \textbf{15.4} & 18.8 & \underline{17.2} & 64.6 & 20.9 & 63.3 & 26.4 & 42.1 & 15.7 & 38.1 & - \\
\midrule
RULER & 8M & 1 & - & - & - & - & - & - & - & - & - & - & - & - & 37.2 & - \\
\midrule
\multicolumn{15}{l}{\textbf{Zero Shot / Reinforcement Fine-tuning}} \\
\midrule
InfiGUI-R1-3B & 32K & - & 51.3 & 12.4 & 44.9 & 7.0 & 33.0 & 14.1 & 58.3 & 20.0 & 65.5 & 28.3 & 43.9 & 12.4 & 35.7 & - \\
\midrule
GUI-G1-3B & 17K & 1 & 50.7 & 10.3 & 36.6 & \underline{11.9} & 39.6 & 9.4 & 61.8 & \underline{30.0} & 67.2 & 32.1 & 32.5 & 10.6 & 37.1 & - \\
\midrule
Qwen2.5-VL-3B & - & - & 31.8 & 4.1 & 32.8 & 4.2 & 24.9 & 4.7 & 43.8 & 12.7 & 42.4 & 15.1 & 17.8 & 2.2 & 22.7 & - \\ \rowcolor{gray!35}
+Decomposed Grounding & - & - & 52.6 & 8.3 & 42.9 & 11.2 & 25.9 & 3.1 & 47.9 & 10.0 & 55.9 & 17.0 & 46.7 & 9.0 & 31.2 & 38.8 \\
Qwen2.5-VL-7B & - & - & 54.5 & 5.5 & 24.7 & 4.2 & 13.7 & 3.1 & 46.5 & 7.3 & 50.8 & 11.3 & 29.9 & 10.1 & 24.5 & - \\ \rowcolor{gray!35}
+Decomposed Grounding & - & - & 60.4 & 13.1 & 33.3 & 8.4 & 27.9 & 6.2 & 50.0 & 13.6 & 63.3 & 17.0 & 51.4 & 16.0 & 33.3 & 42.0 \\
\midrule
GUI-R1-3B & 3K & 9 & 40.9 & 4.8 & 47.8 & 2.8 & 27.9 & 6.3 & 65.3 & 19.1 & 58.2 & 18.9 & 29.0 & 2.2 & 30.9 & - \\ \rowcolor{gray!35}
+Decomposed Grounding & - & - & 63.6 & 13.1 & 55.6 & 4.9 & 31.5 & 6.3 & 61.8 & 16.4 & 62.7 & 20.8 & 44.9 & 10.1 & 37.1 & 46.0 \\
GUI-R1-7B & 3K & 9 & 57.1 & 8.3 & 37.9 & 8.4 & 28.4 & 6.3 & 54.9 & 10.9 & 59.9 & 13.2 & 41.1 & 13.5 & 32.1 & - \\ \rowcolor{gray!35}
+Decomposed Grounding & - & - & \underline{66.9} & 15.2 & 50.0 & 10.5 & 32.5 & 4.7 & 59.0 & 13.6 & 68.4 & 24.5 & \textbf{60.7} & 18.0 & 39.3 & 48.6 \\
\midrule
UI-R1-3B & 136 & 8 & 22.7 & 4.1 & 27.3 & 3.5 & 11.2 & 6.3 & 42.4 & 11.8 & 32.2 & 11.3 & 13.1 & 4.5 & 17.8 & - \\
UI-R1-E & 2K & 8 & 46.1 & 6.9 & 41.9 & 4.2 & 37.1 & 12.5 & 56.9 & 21.8 & 65.0 & 26.4 & 32.7 & 10.1 & 33.5 & - \\ \rowcolor{gray!35}
+Decomposed Grounding & - & - & 63.6 & 17.2 & 59.6 & 10.0 & 43.7 & 6.3 & 66.0 & 21.8 & 68.4 & \underline{43.4} & 56.1 & 19.1 & 43.3 & 52.6 \\
\midrule \rowcolor{gray!35}
\ours-3B & 9K & 2 & 53.2 & 9.0 & 50.5 & 8.4 & 44.2 & \textbf{20.3} & 62.5 & 22.7 & 65.5 & 22.6 & 35.5 & 12.4 & 37.9 & - \\ \rowcolor{gray!35}
+Decomposed Grounding & - & - & \underline{66.9} & 16.6 & 58.1 & \underline{11.9} & 47.2 & 10.9 & \underline{66.7} & 24.5 & 72.3 & 34.0 & 58.9 & \textbf{22.5} & \underline{45.0} & \underline{54.4} \\ \rowcolor{gray!35}
\ours-7B & 9K & 2 & 64.3 & 15.2 & 53.0 & 9.8 & \underline{49.2} & 14.1 & \textbf{72.9} & 25.5 & \underline{75.1} & 30.2 & 45.8 & 20.2 & 44.0 & - \\ \rowcolor{gray!35}
+Decomposed Grounding & - & - & \textbf{79.1} & \textbf{24.1} & \textbf{60.6} & 11.2 & \textbf{53.3} & 10.9 & 66.0 & 26.4 & \textbf{79.1} & 39.6 & \underline{59.8} & \textbf{22.5} & \textbf{48.7} & \textbf{59.2} \\
\bottomrule
\end{tabular}
\label{tab:ScreenSpot_Pro}
\end{table*}

\begin{table}[t]
\centering
\footnotesize
\setlength{\tabcolsep}{3pt}
\caption{Grounding accuracy on ScreenSpot-v2. Bold and underlined Results represent the best performance and the second-best performance, respectively.}
\vspace{-10pt}
\begin{tabular}{lcc cc cc c}
\toprule
\multirow{2}{*}{\textbf{Method}} & \multicolumn{2}{c}{\textbf{Mobile}} & \multicolumn{2}{c}{\textbf{Desktop}} & \multicolumn{2}{c}{\textbf{Web}} & \multirow{2}{*}{\textbf{Avg.}} \\
\cmidrule(lr){2-3} \cmidrule(lr){4-5} \cmidrule(lr){6-7}
& Text & Icon & Text & Icon & Text & Icon & \\
\midrule
SeeClick & 78.4 & 50.7 & 70.1 & 29.3 & 55.2 & 32.5 & 55.1 \\
OS-Atlas-4B & 87.2 & 59.7 & 72.7 & 46.4 & 85.9 & 63.0 & 71.9 \\
OS-Atlas-7B & 95.0 & 73.3 & 92.8 & 64.9 & 89.6 & 72.4 & 83.7 \\
Aguvis-7B & 94.9 & 80.1 & 95.0 & 77.9 & 91.4 & 69.9 & 85.6 \\
Qwen2.5-VL-3B & 96.1 & 74.8 & 87.8 & 53.0 & 86.9 & 70.4 & 80.7 \\
Qwen2.5-VL-7B & 98.4 & 84.8 & 88.4 & 74.7 & 92.5 & 77.6 & 87.5 \\
GUI-R1-3B & 98.1 & 79.0 & 94.0 & 66.7 & 93.3 & 69.2 & 85.2 \\
GUI-R1-7B & 98.8 & 86.4 & 92.3 & 79.4 & 92.1 & 77.2 & 88.7 \\
UI-R1-E & \underline{99.6} & 80.1 & \textbf{95.6} & 75.4 & 91.6 & 81.2 & 88.7 \\
UI-TARS-2B & 95.2 & 79.1 & 90.7 & 68.6 & 87.2 & 78.3 & 84.7 \\
UI-TARS-7B & 96.9 & \underline{89.1} & \underline{95.4} & \underline{85.0} & \underline{93.6} & \underline{85.2} & \underline{91.6} \\
UI-TARS-72B & 94.8 & 86.3 & 91.2 & \textbf{87.9} & 91.5 & \textbf{87.7} & 90.3 \\
\midrule
\ours-3B & \underline{99.6} & 86.4 & 93.9 & 74.5 & 91.8 & 77.6 & 88.6 \\
\ours-7B & \textbf{100.0} & \textbf{91.1} & \textbf{95.6} & 84.8 & \textbf{94.2} & 83.0 & \textbf{92.1} \\
\bottomrule
\end{tabular}
\label{tab:ScreenSpot-v2}
\end{table}

We evaluate the grounding ability on ScreenSpot-v2~\citep{0003WXWSJC0CL025} and ScreenSpot-Pro~\citep{abs-2504-07981}. ScreenSpot-v2 is a corrected version of the original ScreenSpot~\citep{ChengSCX0Z024}, providing evaluation of GUI grounding capability across mobile, desktop, and web platforms. ScreenSpot-Pro focuses on high-resolution professional environments, featuring expert-annotated tasks spanning 23 applications, five industries, and three operating systems. 
Since the images in ScreenSpot-v2 are already pre-cropped while those in ScreenSpot-Pro are full, uncropped displays, we evaluate our decomposed grounding with selection method exclusively on ScreenSpot-Pro. 

\vspace{5pt}
\noindent\textbf{Effectiveness of Inference Enhancement.} 
As shown in Tab.~\ref{tab:ScreenSpot_Pro}, decomposed grounding with selection shows significant improvements on ScreenSpot-Pro. 
It provides a universal and substantial performance boost across all tested models, regardless of their original training paradigm (SFT or RFT). 
For instance, it elevates the average score of OS-Atlas-7B from 18.9 to 33.1 (\textbf{+75.1\%}), and boosts Aguvis-7B from 20.4 to 36.5 (\textbf{+78.9\%}). The accuracy of selection (about 87\%) can be inferred from pass@4 metric in Tab~\ref{tab:ScreenSpot_Pro}.
The consistent improvement proves the effectiveness of decomposed grounding with selection and its high applicability as a plug-and-play inference enhancement.

\vspace{5pt}
\noindent\textbf{Effectiveness of Training Enhancement.} 
Besides, Tab.~\ref{tab:ScreenSpot_Pro} shows that \ours-3B and \ours-7B models, even without decomposed grounding, establish a new state-of-the-art among 3B and 7B models on ScreenSpot-Pro. 
Trained on only 9K examples with 2 epochs, they (37.9 for 3B and 44.0 for 7B) surpass other RFT-based models like UI-R1-E (33.5), InfiGUI-R1-3B and GUI-R1-7B (32.1). 
\ours-7B even outperforms the much larger model UI-TARS-72B (38.1) trained on approximately 50 billion tokens. 
On the ScreenSpot-v2 benchmark (Tab.~\ref{tab:ScreenSpot-v2}), our \ours-7B also achieves state-of-the-art grounding accuracy with an average score of 92.1.
The above results demonstrate the effectiveness of our proposed ``Simple Thinking'' reward, continuous grounding reward, and cropping-based resampling for improving the training of GUI agents.

Overall, as shown in Tab.~\ref{tab:ScreenSpot_Pro}, using both our training and inference enhancements (\ours-7B + Decomposed Grounding) brings 23\% grounding accuracy improvement over the best baseline (JEDI-7B) on ScreenSpot-Pro.

\subsection{Agent Capability Evaluation}

\begin{table}[t]
\centering
\footnotesize
\setlength{\tabcolsep}{1mm}
\caption{Type, GR and SR on AndroidControl-Low and AndroidControl-High. Bold and underlined Results represent the best performance and the second-best performance, respectively.}
\vspace{-10pt}
\begin{tabular}{lcccccc}
\toprule
\multirow{2}{*}{\textbf{Models}} & \multicolumn{3}{c}{AndroidControl-Low} & \multicolumn{3}{c}{AndroidControl-High} \\
\cmidrule(lr){2-4} \cmidrule(lr){5-7} 
 & Type & GR & SR & Type & GR & SR \\
\midrule
Os-Atlas-4B & 64.6 & 71.2 & 40.6 & 49.0 & 49.5 & 22.8 \\
Os-Atlas-7B & 73.0 & 73.4 & 50.9 & 57.4 & 54.9 & 29.8 \\
Qwen2.5-VL-3B & 80.5 & 79.4 & 67.8 & 64.4 & 46.1 & 44.4 \\
Qwen2.5-VL-7B & 78.0 & 87.1 & 68.7 & 69.1 & 59.1 & 50.1 \\
UI-R1-E & \underline{87.0} & 77.8 & 71.4 & 66.4 & 37.8 & 36.9 \\
GUI-R1-3B & 83.7 & 81.6 & 64.4 & 58.0 & 56.2 & 46.5 \\
GUI-R1-7B & 85.2 & 84.0 & 66.5 & 71.6 & \textbf{65.6} & 51.7 \\
\midrule
\ours-3B & 85.4 & \underline{87.6} & \underline{74.3} & \underline{78.6} & 60.7 & \underline{56.9} \\
\ours-7B & \textbf{87.7} & \textbf{88.1} & \textbf{77.6} & \textbf{80.1} & \underline{61.9} & \textbf{60.6} \\
\bottomrule
\end{tabular}
\label{tab:agent-task}
\end{table}

In addition to grounding-specific benchmarks, we also evaluate \ours-3B and \ours-7B on AndroidControl~\citep{LiBLRCTR24} to assess its general agent capabilities. 

Following the evaluation setting of OS-Atlas~\citep{0003WXWSJC0CL025}, we use three metrics: action type prediction accuracy (Type), grounding accuracy (GR), and the overall step success rate (SR). 
Type accuracy measures the exact match for the predicted action (e.g., click or scroll). 
For GR, a prediction is considered successful if it falls within a 14\% screen-width radius of the ground-truth coordinate. 
The SR deems a step successful only if both the action type and all its associated arguments (e.g., coordinates for a click, direction for a scroll, or text for an input) are correct.

Following OS-Atlas, we use 7,708 examples for a fair comparison, while some works (e.g., Aguvis~\citep{abs-2412-04454} and UGround~\citep{GouWZXCS0025}) randomly sample 500 action steps for testing. 
The evaluation is conducted under two distinct settings. 
In AndroidControl-Low, the agent receives a specific, low-level instruction for each step. 
In contrast, AndroidControl-High provides the agent with a high-level goal, requiring it to infer the correct action for the current step based on the conversation history. 
We also note the strong performance of models like InfiGUI-R1~\citep{abs-2504-14239} and Aguvis~\citep{abs-2412-04454} on AndroidControl. However, they leverage much larger datasets (32K/1M) compared to our 9K training samples, and InfiGUI-R1's training data includes AndroidControl.

As depicted in Tab.~\ref{tab:agent-task}, \ours-7B achieves the best performance (SR: 77.6 and 60.6) compared to other RFT models, including UI-R1-E (SR: 71.37 and 35.88), GUI-R1-3B (SR: 64.41 and 46.55) and GUI-R1-7B (SR: 66.52 and 51.67). 
For Type, a similar trend can be observed. 

For GR, \ours archives the best performance in most cases except that GUI-R1-7B shows higher GR (65.6) in the AndroidControl-High setting.
We attribute it to different training philosophies: their prolonged training on a smaller dataset for 9 epochs may foster specialization, whereas our training on a larger, more diverse dataset for 2 epochs prioritizes generalization. 
This hypothesis is supported by our model's dominant performance on dedicated grounding benchmarks ScreenSpot-Pro and ScreenSpot-v2.

Overall, the remarkable results on AndroidControl demonstrate that the improvements gained from our methods are not confined to improving grounding capability but also translate effectively to better decision-making in multi-step agent scenarios. 

\subsection{Ablation Study}
\label{sec:exp_ablation}

\begin{figure}[t]
\centering
\includegraphics[width=1\columnwidth]{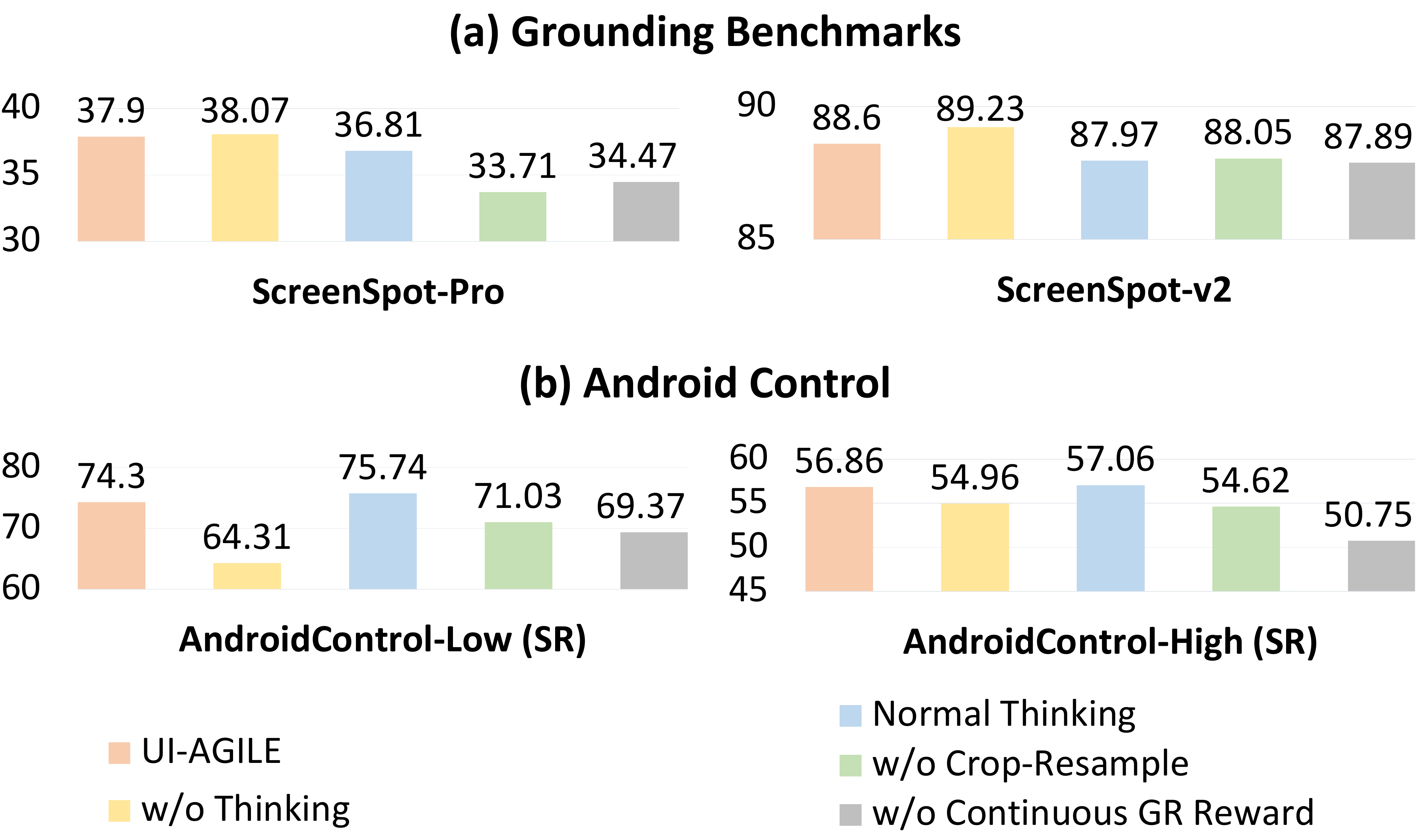} % Reduce the figure size so that it is slightly narrower than the column.
\caption{Ablation result on the three benchmarks.}
\vspace{-10pt}
\label{figure:ablation}
\end{figure}

\noindent\textbf{Ablation Experiments on Training.} 
To verify the contribution of each training technique, we conduct an ablation study of \ours-3B and the results are shown in Fig.~\ref{figure:ablation}(a) and Fig.~\ref{figure:ablation}(b). 
Due to page limit, the performance on AndroidControl w.r.t. Type and GR are provided in Appendix~\ref{sec:ablation_type_gr}.
From the results, we can observe that:
\begin{itemize}
    \item Applying continuous grounding reward and cropping-based resampling improves the performance by 10\% and 12.4\% on ScreenSpot-Pro, respectively. The former incentivizes more precise localization to the target’s center and the later helps avoid ineffective training with zero reward. They also slightly improve grounding accuracy on ScreenSpot-v2 where the performance of base model is already high and it is difficult to achieve significant gains.

    \item Comparing the three strategies (``No Thinking'', ``Simple Thinking'', ``Normal Thinking''), we can observe that:
    \begin{itemize}
        \item Grounding accuracy descends, while Android Control accuracy ascends.
        \item ``No Thinking'' performs significantly worse on the Android Control task.
        \item ``Simple Thinking'' is more accurate in grounding than ``Normal Thinking'' yet achieves comparable performance on Android Control.
        \item ``Normal Thinking'' requires 1.7x more training time than ``Simple Thinking''.
    \end{itemize}
    Hence, we can conclude that ``Simple Thinking'' can achieve the balance between effectiveness and efficiency.
\end{itemize}

\vspace{5pt}
\noindent\textbf{Analysis of Selection Accuracy}
\label{sec:selection_analysis}
For our Decomposed Grounding with Selection method, the selection accuracy, inferred from the pass@4 metric in Tab.~\ref{tab:ScreenSpot_Pro}, is 82.26\%. We attempted to fine-tune the selection VLM, teaching the model to output ``Yes'' if the intruction matches the element otherwise ``No''. The hard negative samples, which are instruction-element pairs, are genetated from the grounding model's own predictions when using our Decomposed Grounding with Selection method. However, this fine-tuning dropped selection accuracy to 68\%, likely due to a mismatch between the training task and the score-based ranking task during inference.

We then explored an alternative: training the VLM to output the index of the correct element image from four candidates given the instruction. This approach perfectly aligns the training and inference tasks. Training on 9k data (identical to the dataset used to train UI-AGILE) improved selection accuracy from ~70\% to 81\%.

\subsection{Analysis of Attempts Per Step}

\begin{figure}[t]
\centering
\includegraphics[width=0.75\columnwidth]{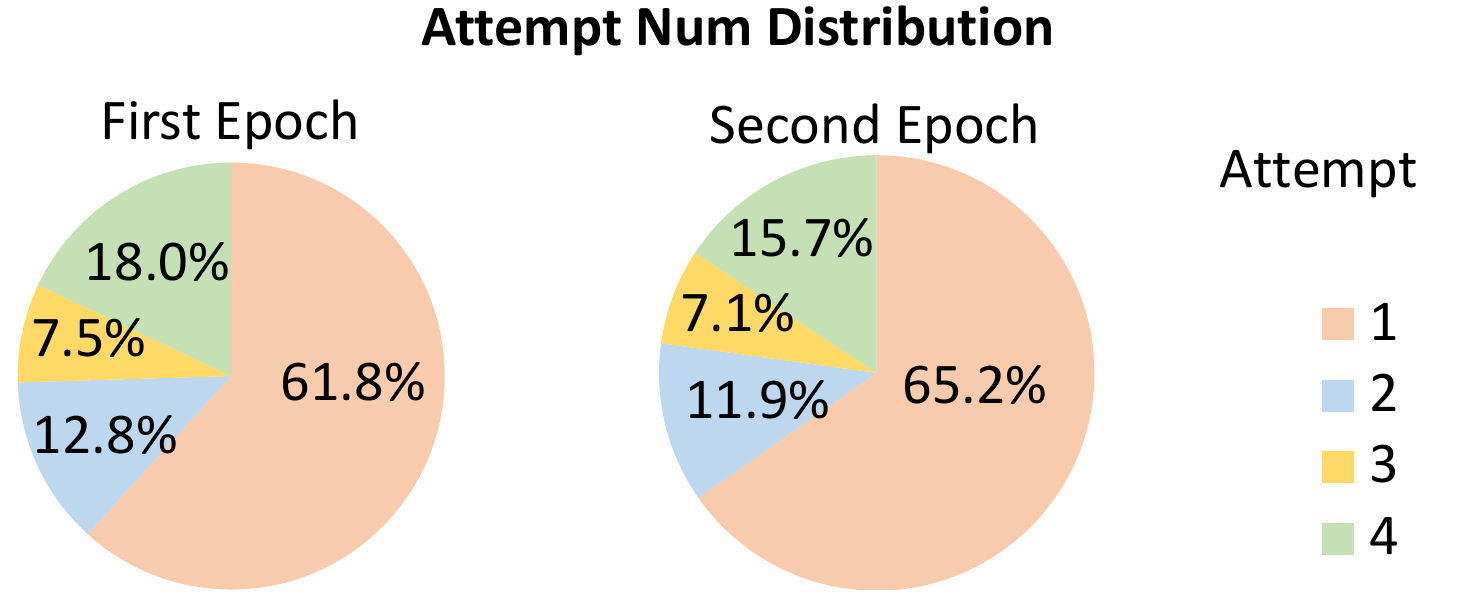} % Reduce the figure size so that it is slightly narrower than the column.
\caption{Distribution of attempts per step throughout the training process. Max attempt number is set to 4.}
\vspace{-10pt}
\label{fig:attempt}
\end{figure}

Fig.~\ref{fig:attempt} shows the distribution of attempts per GRPO training step using cropping-based resampling strategy, where each step processes a batch of two training samples. In the first epoch, we find that only 61.8\% of training steps are fully successful on the initial attempt (i.e., both samples in the batch are solved without resampling). This means that without our strategy, 19.1\% to 38.2\% of training samples would have provided no learning signal. Overall attempt numbers decrease in the second epoch, demonstrating that the model learns from the samples salvaged by our cropping-based resampling. 

%% file: tex/related_work.tex
\section{Related Work}
\label{sec:related}

\vspace{5pt}
\noindent\textbf{Reinforcement Learning for Large Models.} 
Recently, RL algorithm, including PPO~\citep{SchulmanWDRK17}, DPO~\citep{RafailovSMMEF23} and GRPO~\citep{abs-2402-03300}, for training large models has gained significant momentum. 
These algorithms have achieved remarkable success in enhancing the reasoning capabilities of large models on complex tasks, with models like OpenAI O1~\citep{abs-2412-16720} and DeepSeek-R1~\citep{abs-2501-12948}.
% setting new standards in mathematics and code generation. 
The efficacy of these approaches promotes their rapid extension into the multimodal domain~\citep{chen2025r1v, abs-2503-07523, abs-2503-01785, abs-2503-07536, abs-2503-07365, abs-2601-01984, abs-2512-19554}.

\vspace{5pt}
\noindent\textbf{GUI Agents.} 
Techniques for GUI agents have evolved rapidly~\citep{abs-2411-04890, abs-2504-20464} in recent years.
Following early works like CogAgent~\citep{HongWLXYJWWD0024} and SeeClick~\citep{ChengSCX0Z024}, most studies trained models to output actions (e.g., clicking with coordinate parameters, keyboard input) based on visual inputs screenshots and user instructions. 
Show-UI~\citep{LinLGYWBLWS25} innovates on visual processing efficiency. OS-Atlas~\citep{0003WXWSJC0CL025}, UGround~\citep{GouWZXCS0025} and Aria-UI~\citep{abs-2412-16256} propose novel, large-scale pipelines to collect and synthesize millions of GUI agent trajectories, significantly improving model generalization.
Aguvis~\citep{abs-2412-04454} introduced a two-stage training process that explicitly uses VLM-generated Chain-of-Thought (CoT) data to teach planning and reasoning. JEDI~\citep{abs-2505-13227} constructs a refusal part by mismatching existing instructions with unrelated screenshots.
Standing out in complexity and scale, UI-TARS~\citep{abs-2501-12326} utilizes the largest dataset and the most intricate training pipeline, which involves SFT and DPO on human annotated CoT data to improve performance.
This data-intensive scaling has motivated a shift towards more effective Reinforcement Learning with Verifiable Rewards paradigms, first explored by UI-R1~\citep{abs-2503-21620} and GUI-R1~\citep{abs-2504-10458}.
InfiGUI-R1~\citep{abs-2504-14239} employs Spatial Reasoning Distillation to enhance cross-modal spatial reasoning capabilities and uses RL to refine the basic reasoner into a deliberative one.
GUI-G1~\citep{abs-2505-15810} leverages Hit-based reward and IoU-based reward for improving GUI agents. RULER~\citep{abs-2510-03230} uses tokens as explicit coordinate markers, letting the model reference positions similar to
gridlines.

%% file: tex/conclusion.tex
\section{Conclusions}
\label{sec:con}

In this paper, we introduce \ours, a comprehensive framework designed to enhance GUI agents' training and inference. 
It tackles the practical challenges of the reasoning-grounding dilemma, ineffective reward, and visual noise.
% During training, our solution integrates three key innovations: a ``Simple Thinking'' reward to foster efficient yet effective reasoning, a continuous grounding reward that incentivizes high-precision localization, and a cropping-based resampling strategy to overcome the sparse reward problem. 
% For inference, we introduce decomposed grounding with selection, a novel method that reduces visual noise and dramatically improves grounding accuracy on high-resolution screens while the inference cost is only slightly increased.
Experimental results demonstrate the effectiveness of our proposed techniques on enhancing GUI agents. Future work may include exploring alternative selection stage methods, such as computing text-element image embedding similarity and cropping as a function call. 

% \paragraph{Limitations and future work.} 
% Despite the promising performance of \ours, the VLM used in its decomposed grounding with selection method is a general-purpose, pre-trained model that is not originally tuned for the selection task. 
% A promising future direction would be to fine-tune this adjudicator model on a curated dataset of candidate UI elements, enhancing its selection accuracy to achieve further gains in overall grounding performance. 

%% file: tex/X_suppl.tex
\clearpage
\setcounter{page}{1}
\maketitlesupplementary
\appendix

\section{Environment and Implementation Details}
\label{sec:env_impl_details}

In the following, we provide the details of experiment environment and our implementation.

\vspace{5pt}
\noindent\textbf{Environment.} 
We use a machine with two Intel(R) Xeon(R) Silver 4314 CPU @ 2.40GHz, 512GB main memory and eight NVIDIA A800 GPU for experiments.

% \ref{vonwerra2022trl}

\vspace{5pt}
\noindent\textbf{Training Details.} We use the trl framework\footnote{\url{https://github.com/huggingface/trl}}
to implement the cropping-based resampling strategy and reward functions. The sampling process is attempted 4 times at most and is bypassed entirely if the bbox's dimensions exceed the target crop size. 
% prior works~\citep{abs-2503-21620, abs-2504-10458, abs-2504-14239, abs-2505-15810}
Following prior works, we use Qwen2.5-VL-3B\footnote{\url{https://huggingface.co/Qwen/Qwen2.5-VL-3B-Instruct}}
and Qwen2.5-VL-7B\footnote{\url{https://huggingface.co/Qwen/Qwen2.5-VL-7B-Instruct}}
as base models.

\begin{figure}[t]
\centering
\includegraphics[width=0.45\textwidth]{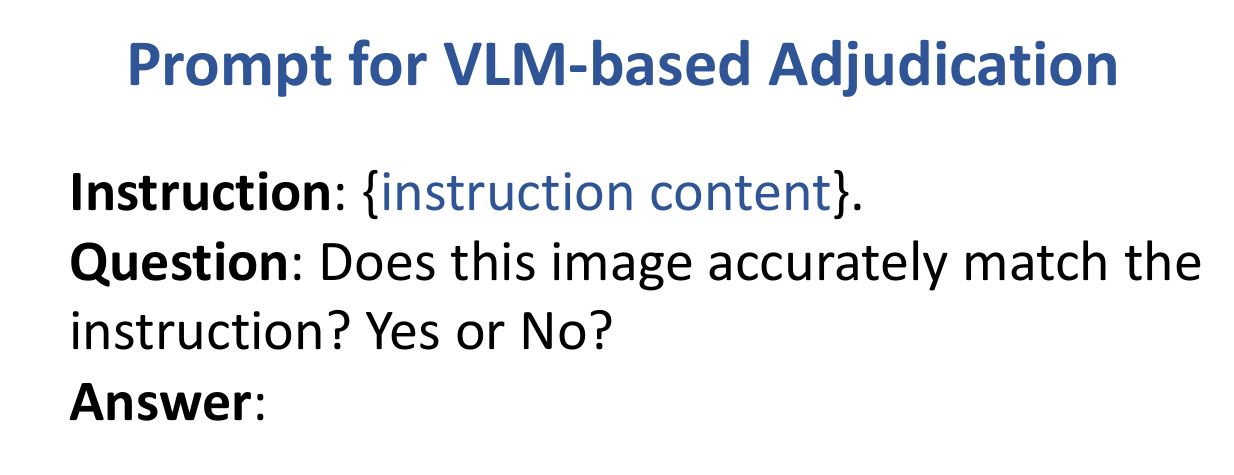}
\vspace{-10pt}
\caption{Prompt for VLM-based adjudication.}
\label{adjudication_prompt}
\end{figure}

\vspace{5pt}
\noindent\textbf{Inference Details.} 
For decomposed grounding with selection, the input image is divided into four sub-images scaling to 60\% of the original dimensions, with adjacent sub-images overlapping by 10\% of the original image's width and height.  
In the element image extraction stage, we define the element's area by creating a simple bounding box centered on the predicted point with the width and height equal to 14\% of the sub-image's width and height. 
We have also explored a more sophisticated approach using OmniParser to refine this bounding box. 
However, it does not improve performance and increases the inference overhead.
% we retained the simpler and more efficient method for our final implementation. 
In the selection stage, we use Qwen2.5VL-7B-instruct
%\footnotemark[1] 
to choose the final answer and the prompt is listed in Fig.~\ref{adjudication_prompt}.

\vspace{5pt}
\noindent\textbf{Hyperparameters.} 
Tab.~\ref{tab:training_hyperparameter} provides the training hyperparameters of \ours where cropping factor is the width and height ratio of new attempted image and last attempted image. 
The hyperparameters in Eq.~\ref{eq:simple_thinking_reward} and Eq.~\ref{eq:continuous_reward} are optimized using a grid search.
Tab.~\ref{tab:eq_hyperparameter} shows the specific values.

\section{Code}
\label{sec:code}

We provide \textbf{the code for our RFT training} and \textbf{the Decomposed Grounding with Selection method} in two separate modules. To avoid potential dependency conflicts, each module is designed to be run in its own conda environment.

To ensure a fair and comprehensive comparison, we conducted extensive experiments on the ScreenSpot-Pro benchmark. This involved re-implementing baseline models and evaluating them with our Decomposed Grounding with Selection method.

% You may provide detailed mathematical derivations, proofs, or other technical details here.

We use the parquet format to store test data in order to reduce the I/O read overhead. 

We still have room for improvement in the implementation of Decomposed Grounding with Selection for Inference, including multi-threading to accelerate processing images and other non-GPU operations.
% \subsection{Pseudocode}

\begin{table}[t]
\centering
\caption{Training hyperparameters.}
\vspace{-10pt}
\begin{tabular}{@{}lc@{}}
\toprule
\textbf{Hyperparameter}     & \textbf{Value} \\ \midrule
learning rate           & from 1e-06 to 4.36e-10 \\
num generations           & 8              \\
num train epochs          & 2              \\
per device train batch size & 4              \\
gradient accumulation steps & 4              \\
cropping factor                & 0.6            \\
sampling attempt num             & 4        \\
\bottomrule
\end{tabular}
\label{tab:training_hyperparameter}
\end{table}

\begin{table}[t]
\centering
\caption{Hyperparameters for Eq.~\ref{eq:simple_thinking_reward} and Eq.~\ref{eq:continuous_reward}}
\vspace{-10pt}
\begin{tabular}{@{}lc@{}}
\toprule
\textbf{Hyperparameter}     & \textbf{Value} \\ \midrule
$l_{\text{ideal\_start}}$           & 120 chars \\
$l_{\text{ideal\_end}}$         & 200 chars              \\
$l_{\text{min}}$          & 50 chars              \\
$l_{\text{max}}$          & 300 chars             \\
$C$                       & 4 \\
\bottomrule
\end{tabular}
\label{tab:eq_hyperparameter}
\end{table}

\section{Additional Results for Ablation Study}
\label{sec:ablation_type_gr}
% Due to page limitations, our ablation study on the AndroidControl benchmark focuses on the most critical metric, the success rate (SR), in the main paper. A detailed breakdown of the other metrics, including grounding accuracy and type prediction accuracy, is provided in Figure~\ref{ablation_type} and Figure~\ref{ablation_acgr}.

\begin{figure}[t]
\centering
\includegraphics[width=0.45\textwidth]{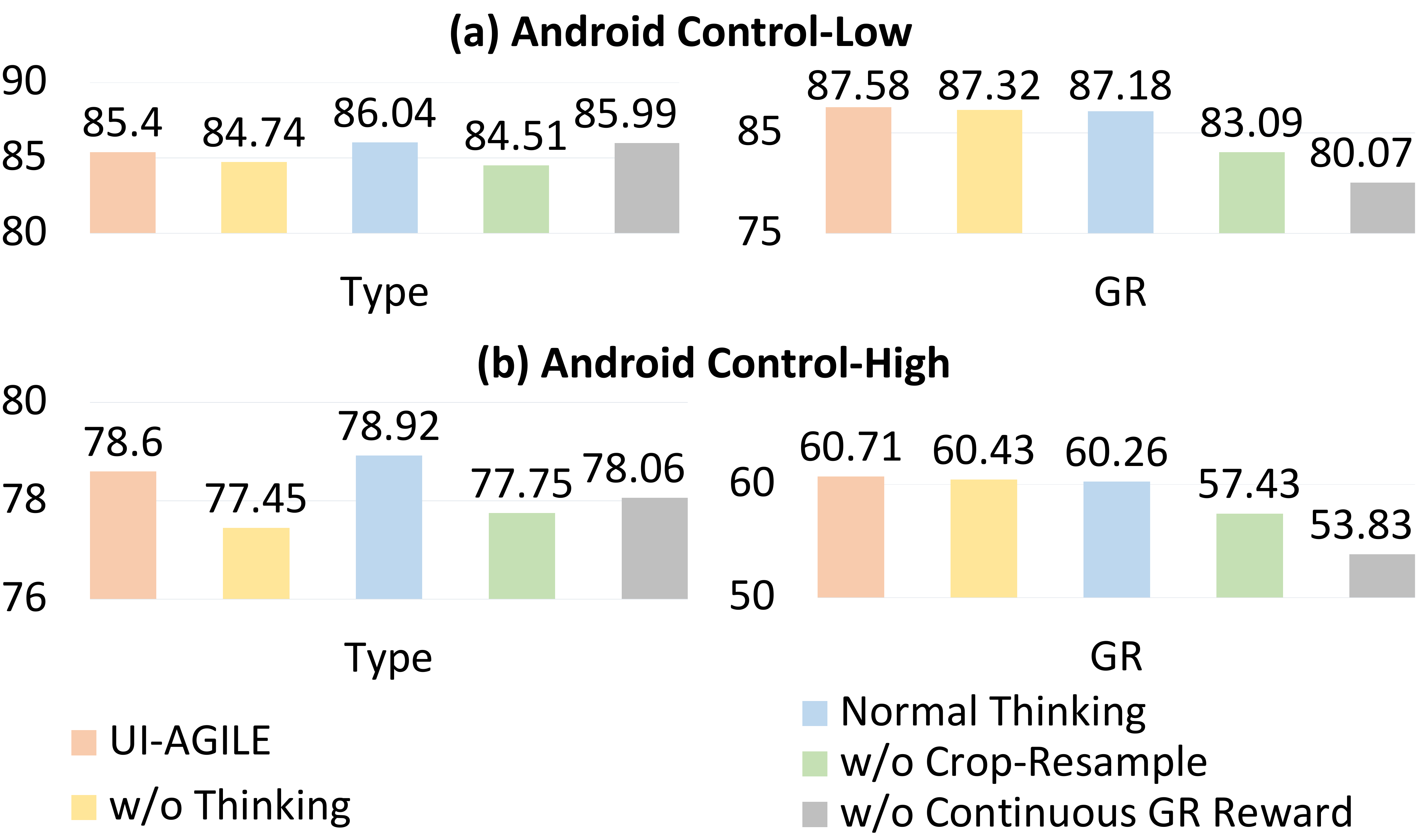}
\caption{Ablation study using Type and GR as evaluation metrics on AndroidControl.}
\label{more_ablation}
\end{figure}

Fig.~\ref{more_ablation} provides results of the ablation study using Type and GR as evaluation metrics on AndroidControl. We can observe that each component in \ours indeed contributes to the overall performance.

\section{Analysis of Inference Time}
\label{sec:inference_time}

% To provide a concrete measure of the computational overhead, 
We report the inference time of our decomposed grounding with selection method on the full ScreenSpot-Pro dataset~\citep{abs-2504-07981} using the vLLM framework~\citep{KwonLZ0ZY0ZS23} and one 80G A800 GPU card.  

% without our method

As a baseline, the standard grounding approach applied to \ours-7B completes the benchmark in \textbf{30 minutes}. 
When applying our method, the decomposed grounding stage takes \textbf{35 minutes}. 
The subsequent VLM-based selection stage requires additional \textbf{4 minutes}. 
The modest increase in overhead is a practical trade-off for the substantial gain of grounding accuracy brought by our method. 